\documentclass[preprint,12pt]{elsarticle}
\usepackage{amssymb}
\journal{Arxiv.org}

\begin{document}

\begin{frontmatter}

\title{  Logical analysis of natural language semantics to solve the problem of  computer understanding}

\author{Yuriy Ostapov}
\address{Institute of Cybernetics of NAS of Ukraine, pr. Acad. Glushkova, 40, Kiev, 03680, Ukraine.
E-mail: yugo.ost@gmail.com }

\begin{abstract}

An object--oriented approach to create a natural language understanding system  is considered.           
The understanding program is a formal system built on the base of predicative calculus.
Horn's clauses are used as well--formed formulas. An inference is based on the principle of resolution. Sentences of natural language are represented  in the view of typical predicate set.
These predicates describe physical objects and processes, abstract objects, categories and semantic relations between objects.
Predicates for concrete assertions are saved in a database. To describe the semantics of classes for physical objects, abstract concepts and processes, a knowledge base is applied.
The proposed representation of natural language sentences  is a semantic net. Nodes of such net are typical predicates.
This approach is perspective as, firstly, such typification of nodes facilitates essentially  forming of processing algorithms and object descriptions,
secondly, the effectiveness of algorithms is increased ( particularly for the great number of nodes), thirdly, to describe the semantics of words, encyclopedic knowledge is used,
and this permits  essentially to extend the class of solved problems.
\end{abstract}

\end{frontmatter}

\section{Introduction}

The given paper is devoted to logical aspects of object--oriented approach to create a natural language understanding system.
This approach was proposed in our works \cite{Ostapov1, Ostapov2, Ostapov3}.

Consider the next tasks connected with computer understanding:

\begin{itemize}
\item  what  the computer understanding means;
\item  how to realize the computer understanding;
\item  how to check the accuracy of computer understanding.
\end{itemize} 

An understanding of complete text is a thorough insight into the sense of assertions which are included in this text.
As regards the computer understanding, it is realized with algorithms to represent the semantics of sentences  and to check such representation.

Consider at first what the assertion  {\it the sense of sentence is ...}  means :\\[10pt]
\hspace*{20pt} 1. It should be distinguished the sense for an author of information from the sense for a reader of this information  as this depends on the volume of knowledge. 
Ontological and epistemological representations are as well of great importance\footnote{For example, consider the sentence {\it Napoleon  lost the battle of Waterloo}. The name {\it Napoleon} can consider as the name of historical person or the name of horse. 
At last, it can assume the migration of souls. Such distinction in semantic representation for the name {\it Napoleon} implies the distinction in the sense of sentence. }.\\
\hspace*{20pt} 2. The sense of sentence is contained, firstly, in the meaning of words, secondly, in the functional role of these words in the sentence, 
thirdly, in properties of objects which are known from previous sentences \footnote{ For instance, a sentence describes actions of {\it Peter}.
It is known  from previous sentences that {\it Peter} is unmarried. This can explain his further actions.}. \\
\hspace*{20pt} 3. For abstract concepts and classes of physical objects, the sense of words is concluded in the standard understanding (in explanatory dictionaries).
For individual objects, the sense consists in properties of these objects. Some words have not the standard meaning ({\it good, beauty, love}), then
it should be took into account an emotional relation of author to their using.\\
\hspace*{20pt} 4.  When we speak about the functional role of words in a sentence, we imply such categories as a subject, objects of influence, an action or internal change of subject, 
other facts of the case (place, time, cause, purpose, tool and method of action). It should be differed an object and its  properties.\\
\hspace*{20pt} 5. A sentence can not be considered separately, but only in the conjunction with other sentences and knowledge of object domain.
The sense of sentences  has as well an implicit character which is contained in actions and events following from actual information of sentences.\\

By this means, the sense of sentences have objective and subjective aspects. It can point to three aspects \cite{Coboz}:

\begin{itemize}
\item  {\it  cognitive (propositional)} - the description of facts, as well as general assertions following from facts;
\item  {\it  communicative} - the organization of text so that the accent is done on certain factors of content;
\item  {\it  pragmatical (illocutive)} - the explanation of sentence pronouncing as the action that is executed with certain purpose.
\end{itemize} 

We limit ourselves only to cognitive  (propositional) aspect. It is necessary to extract an objective component in the sense of declarative sentences.
This component presents the general content for a software engineer and an expert which checks the accuracy of understanding program work.
To extract this  objective component, we assume that, firstly, both are based on certain ontology and epistemology, 
secondly, meanings of words are understood equally (according to an explanatory dictionary), thirdly, both follow standard rules of grammar 
(for English see, for example, \cite{Quirk}).

To answer the question {\it How to realize the computer understanding?},  we must (as noted above) be based on certain ontology and epistemology.
As such basis, we use {\it modern empiricism}  \cite{RusselB}. Empiricism is founded on modern physical science concerning the structure of world and explains
the reliability of scientific conclusions. We take the main thesis of empiricism: {\it all synthetic knowledge is based on experience}.

The world consists of {\it  physical objects}, which correspond to certain domains of {\it  physical place} in a given point of time.  
Physical objects can act or change  their state. These actions of objects and processes of internal state change (events)  are named  {\it facts}.
By this means, {\it the real world consists of facts}. As B.Russell points, {\it physical facts, for most part, are independent not only of our volitions,
but even of our existence} \cite [part 2, ch.11]{RusselB}. We describe both real (or possible) facts and general assertions with the help of declarative sentences.  
The truth of concrete sentences (for real actions and events) consists in the conformity with available facts. 
For our approach, the sentence of  natural language is presented with a set of predicates describing {\it actions, events, 
states, processes, persons, organizations, things, machines}, and other classes of essences and connections.

To answer the question {\it How to check the accuracy of computer understanding?}, it should be took into account the above determination of sense.
To establish the conformity of sentence sense and its semantic representation, it will be sufficient to use simple and logical questions. 
Simple questions permit to check the availability of all categories (subject, objects, etc) in representations of facts.
Logical questions examine the ability of computer system to do inferences on the basis of present facts and knowledge.

\section{Levels of computer understanding}

Preparatory proceeding to the computer understanding, it should be specified  basic principles of computer work: \\[10pt]
\hspace*{20pt} 1.A computer has  a timepiece. This denotes that the computer {\it measures a physical time}.\\
\hspace*{20pt} 2. A computer permits to save numbers and symbols. Another way, the computer {\it has a memory}. On--line storage consists of cells.
Each cell can save a symbol  and has a certain address. On--line storage is short--term memory. Long--term memory is realized using peripheral.\\
\hspace*{20pt} 3. A computer maps points on a display. Consequently, the computer {\it forms points of discrete space with two coordinates}.\\
\hspace*{20pt} 4. A computer executes arithmetic. Then, it can say that the computer {\it understands what arithmetic mean}.\\
\hspace*{20pt} 5. A computer moves a symbol from one address to other. This permits {\it to form abstract words as sequences of symbols}.\\
\hspace*{20pt} 6. A computer executes a program, which consists of a sequence of commands. Each command contains the code of operation and addresses. 
Commands execute arithmetic, transfer of symbols,  and  conditional transfer. Conditional transfer changes the sequence of command execution.\\
\hspace*{20pt} 7. A computer can "see", "hear" by means of peripheral. This means that the computer {\it has perceptions}.\\
\hspace*{20pt} 8. A computer (robot) can go in physical space  and  implement various physical  actions using special mechanisms.\\
\hspace*{20pt} 9. As an operation system  realizes different actions, controls programs, remembers all actions and events, it can speak 
about {\it computer consciousness}.\\

The construction of computer permitted  A.Turing to express the hypothesis about the ability of computer to answer questions so that 
an answer of a person and the computer can not be distinguished. Another way, the computer can understand questions and adequately answer.

It can distinguish four levels of computer understanding.

At the first level we take into account  only such grammatical categories as subject, objects, place and time action (event), as well as other adverbial modifiers in a sentence. 
This approach is realized with syntax analysis. For this level a computer system can answer simple questions. 
Such approach is of frequent use to translate from one language to other by means of computer.

At the second level paradigmatic and syntagmatic relations between words are established. 
Sentences are presented with semantic net where nodes have a simple structure. Investigations in \cite{Apr,Boris,Leont,Melchuk, Riesbeck,Schank,IPP,Birn,Self} relate to this level.
At the second level an understanding program can answer many questions using an inference. The understanding of semantics (for actions, events and physical objects) is not sufficient and 
is contained usually in inference rules.

At the third level sentences are represented with a set of predicates (the object--oriented approach)\cite{Ostapov1,Ostapov2,Ostapov3}. Properties and changes of physical and abstract object are described with typical predicates\footnote{Our approach is ontological.
Based on functions, properties and structure of real objects, we advert to  the description of words. 
If  we shall form a special structure (frame) for every word, then we shall be at a deadlock as the variety of words is enormous. To solve this problem, we use encyclopedic knowledge.}. 
Physical, physiological, and psychological characteristics are estimated using special scales. If measuring is executed with a physical device, then the semantics of property is described by physical regularities.
For psychological characteristics, it is necessary to use algorithms of recognition. To determine economical activities, special algorithms are demanded.

At third level there is the certain advance to describe actions and processes. In the work \cite{Schank} the classification of action was proposed.
We use it in the modified view:\\[10pt] 
{\it  PROCESS} --- long purposeful occupation ({\it job, sport, studies});\\
{\it  PROPEL} --- applying a force to an object;   \\
{\it  MOVE} ---  movement of body part; \\
{\it  INGEST} ---  ingesting something inside;\\
{\it  EXPEL} ---  expelling something from a subject;\\
{\it  GRASP} ---  grasping an object;\\
{\it  GO} --- displacement of subject;\\
{\it  TRANSFER} ---  change of general relation for a subject ({\it to buy, to sell, to come into fortune});\\
{\it  ATTEND} ---  perceptions of subject ({\it to see, to hear, to touch});\\
{\it  MESSAGE} ---  transmission of information between a subject and object;\\
{\it  BE} ---  identity of subject and object, existence of subject or connection between a subject and  certain class of objects;\\
{\it  CHANGE} --- transition of subject to another internal state (event);\\
{\it  CREATE} ---  thinking ({\it decision-making, problem-solving, prediction});\\
{\it  HAVE} ---  possession of object.\\

Clearly this classification is not sufficient as it does not take account of semantics for many verbs. To describe the semantics of simple movements of hand and leg, it can use cinematic equations.
The complex action (for example, {\it to lead an orchestra, to play tennis, to send a message}) is a sequence of simple movements and described with {\it verbal definition} and {\it operation} (see further).
T.Winograd was a pioneer in this direction. In the work \cite{Winograd} virtual movements of robot are realized by certain programs.

To represent concrete and general sentences (for a given object domain), we apply a semantic net.  Nodes of such net are typical predicates.
This approach is perspective ( particularly for the great number of nodes). Firstly, such typification of nodes simplifies greatly  forming  of processing algorithms and object descriptions,
secondly, the  effectiveness of algorithms is increased, thirdly,  we apply encyclopedic knowledge to describe the semantics of words,
and this  extends essentially the class of solved problems.

At the fourth level the object--oriented approach is connected with functions of perception and physical movement.
To do this, it is necessary to create a robot which has  organs of sense, executes physical measurements, as well as realizes physical movements.
The availability  of recognition algorithms permits to teach such robot the cognizance of physical objects and determination of their properties.
Then it can use {\it visual determinations} to teach basic words \cite [part II-ch.2]{RusselB}. Other words are determined {\it verbally}. 
Besides, such robot can implement certain physical actions. This speaks about more deep understanding of these actions.

Searle's point is that {\it the mere carrying out of a successful algorithm  does not in itself imply that any understanding has taken place} \cite[ch.1]{Penrose}. 
However, at the fourth level a robot and a person overcome the same path to the understanding of that how to represent physical objects and actions (events)  associated with words 
(using the realization of physical actions and recognition algorithms of objects).
Besides, the robot recognizes words of certain language and discovers their grammatical role in a sentence, takes account of sense relations between words, checks the accuracy of word combinations.
Consequently, at the fourth level there is much common in computer and human understanding although physical principles of work are different. 
No all actions of  robot can be described formally, that is, it can not be spoken that the robot works only according to strict algorithms.

Thus, the fourth level permits essentially to approximate to the human understanding. However, there are open problems connected with the understanding of such abstract concepts as {\it good,  beauty, love, truth}, etc.
At the moment it is not clear how to do this. Some questions can be solved now. For example, the program of understanding can be taught to create new classes using  algorithms of classification.
We can form new properties by means of analogous algorithms. Induction and analogy can be used to solve  creative tasks \cite {Polya}.

\section{Typical predicates to represent sentences}

To represent  concrete and general sentences, it is necessary to use the next predicates for:

\begin{itemize}
\item  physical objects (persons, things, machines, animals, natural phenomena);
\item  actions and processes connected with physical objects;
\item  events and states for physical objects;
\item  categories connected with objects, actions and processes (time, place, property);
\item  abstract objects;
\item  sense relations between actions, processes, objects, events and states.   
\end{itemize}

Physical objects are described with predicates of type $\psi(u_{1},u_{2}, ...,u_{n})$ where $u_{1},u_{2}, ...,u_{n}$ are variables
providing, in the aggregate, the identification of object. For example, it is necessary to use as such variables  for persons: {\it first} and {\it second name, sex, age, place of birth}, etc.
A concrete object is presented by means of substitution of individual constants instead variables. 
Each real object is characterized by a compact connected domain of {\it physical space}  at a given point of time.
Points of this domain have coordinates $(x_{1},x_{2},x_{3})$, as well as physical parameters (mass, charge, temperature, etc)\cite [App.A]{Carnap}. 
The predicate $\psi$ is true during a certain interval of time if a real object exists in this interval.

Categories connected with objects, actions and processes  are presented with predicates of type $\delta(y_{1},y_{2}, ...,y_{n})$ where $y_{1},y_{2}, ...,y_{n}$ are variables
providing, in the aggregate, the identification of given category. As categories, we consider a time, place, and property of object. 
Concrete category is formed by means of substitution of individual constants instead variables. Every real time is characterized by an interval of {\it physical time},
the place corresponds to a domain of {\it physical space} with coordinates $(x_{1},x_{2},x_{3})$. The property of physical object is a characteristic  conditioned by the physical and psychological regularities.

Sense relations between objects are described with predicates of type $\mu(q_{1},q_{2}, ...,q_{n})$ where $q_{1},q_{2}, ...,q_{n}$ are variables.
Abstract objects correspond to predicates of type $\rho(r_{1},r_{2}, ...,r_{n})$ where $r_{1},r_{2}, ...,r_{n}$ are variables.
A concrete sense relation (abstract object)  is formed using the substitution of individual constants instead variables.

Actions (processes) connected with physical objects  are presented with predicates of type $\varphi(v_{1},v_{2}, ...,v_{n})$ where $v_{1},v_{2}, ...,v_{n}$ are variables
providing, in the aggregate, the identification of given action or process. As such variables, we use a subject, objects of influence, time and place of action, as well as other facts of the case.
As these factors correspond to other predicates, to point to them, we use unique codes appointed these predicates. A concrete action (process) is described using the substitution of individual constants instead variables. 
A real action (process) is characterized by a set of compact connected domains in {\it a physical space--time}\footnote {Hereafter we are dealing with classical (Galilean) space--time\cite [ch.5]{Penrose}.
For the object domain connected with social system, it is sufficient.}. Each such domain consists of points  with coordinates $(x_{1},x_{2},x_{3}, t)$ and physical parameters.
The predicate $\varphi$ is true in a certain interval of time if a set of domains corresponds to a real (i.e., checked) fact. 
A place of action (process) embraces domains of subject, objects of influence in physical space with coordinates $(x_{1},x_{2},x_{3})$ during a given interval of time.

Events (states) connected with physical objects are presented with predicates of type $\gamma(w_{1},w_{2}, ...,w_{n})$ where $w_{1},w_{2}, ...,w_{n}$ are variables
providing, in the aggregate, the identification of this event and state. As such variables, we use codes of subject, objects of influence, time and place, etc.
A concrete event (state) is described using the substitution of individual constants instead variables. 
Each real event (state) is characterized by a set of compact connected domains in {\it a physical space--time}. Each  domain contains points  with coordinates $(x_{1},x_{2},x_{3}, t)$ and physical parameters.
The predicate $\gamma$ is true in a certain interval of time if a  set of domains corresponds to a real fact. 
A place of event (state) embraces domains of subject, objects of influence in physical space with coordinates $(x_{1},x_{2},x_{3})$ during a given interval of time.

Go on now to the detailed description of above types of predicates. Only after that  we can consider the formal representation for concrete and general sentences.
Values of variables, as a rule, correspond to names of variables, otherwise these values are explained. If we speak about the type of something, this means that 
values of variables are divided into several classes. As an illustration, we can point to some names of classes.  Codes of objects, action, events and states included in descriptions of predicates pose references to these objects, actions, events and states.
These references present semantic relations between  objects, actions, events and states.

\subsection{Typical predicates of physical objects}

Predicates of this type are divided into the next sorts: {\it  person, organization, thing, machine, animal, nature}.
The variable {\it  number} in these predicates denotes the grammatical number of object in a sentence: plural or singular.

\subsubsection{Description of persons}

A corresponding predicate takes the form:\\[10pt] 
{\it  person (cod\_pers, sex,  age, number, first\_name, second\_name, cod\_pl\_birth, cod\_dt\_birth, nath, lang, face, nose, eyes, hair, stature, prof)}\\

\noindent where {\it  cod\_pers} is a code of person, {\it  cod\_pl\_birth} is a code of birth place, {\it cod\_dt\_birth} is a code of birth  date, {\it nath} is  a nationality, {\it  lang} is  a mother tongue, {\it  prof} is a profession.

\subsubsection{Description of organizations}

A  predicate of organization is of the form:\\[10pt] 
{\it  organization (cod\_org, name\_org,  typ\_org , number, cod\_loc\_org, director) }\\

\noindent where {\it cod\_org}  is a code of organization, {\it name\_org} is a name of organization,  {\it typ\_org}  is a type of organization, {\it cod\_loc\_org} is a code of organization location.

\subsubsection{Description of things}

A corresponding predicate is of the form:\\[10pt] 
{\it  thing (cod\_th, name,  maj\_class, number, weight, color, length, height, thickness, cod\_owner) }\\

\noindent where {\it  cod\_th} is a code of thing, {\it  maj\_class} is a name of major class.

\subsubsection{Description of machines}

A machine differs from a thing by the availability of energy source and transformation of one energy to other. 
A predicate of machine takes the form:\\[10pt] 
{\it machine (cod\_mach, name,  function, number, typ\_eng, color, trademark, name\_prod, cod\_owner)  }\\

\noindent where {\it  cod\_mach} is a code of machine, {\it  typ\_eng} is a type of engine, {\it  name\_prod} is a name of producer.

\subsubsection{Description of animals}

A predicate of animal takes the form:\\[10pt] 
{\it  animal (cod\_an, typ\_an, number, maj\_class, weight, name, color,  cod\_owner)}\\

\noindent where {\it  cod\_an} is a code of animal, {\it   typ\_an} is a type of animal, {\it  maj\_class} is a name of major class.

\subsubsection{Description of natural phenomena}

A corresponding predicate is of the form:\\[10pt] 
{\it  nature(cod\_nat,  typ\_nat, number, name, charact)   }\\

\noindent where {\it   cod\_nat} is a code of phenomenon, {\it   typ\_nat} is a type of phenomenon, {\it   charact} is a characteristic of phenomenon.

\subsection{Typical predicates of categories}

Predicates of this type are divided into the next sorts: {\it  place, time, property}.

\subsubsection{Description of place}

A predicate of place takes the form:\\[10pt] 
{\it   place (cod\_pl, country,  typ\_reg, name\_reg, ter\_entity, name\_ter, locat, name\_loc, constr, name\_constr, add\_inf\_constr, fin\_locat, name\_room)}\\

\noindent where {\it  cod\_pl} is a code of place, {\it  typ\_reg} is a type of region, {\it  name\_reg} is a name of region, {\it  ter\_entity} denotes a territorial entity ({\it  town, village}), {\it name\_ter} is a name of territorial entity,
{\it locat} is a location ({\it street, square, park, line}), {\it name\_loc} is a name of location, {\it constr} is a construction ({\it house, theatre, station, industrial object}), {\it name\_constr} is a name of construction,
{\it  add\_inf\_constr} is additional information for construction ({\it  stair, roof, garret, floor}), {\it  fin\_locat} is the designation of final location ({\it  apartment, hall, library, office, restaurant, cafe}),
{\it  name\_room} is the designation of room ({\it  living room, kitchen, bathroom, bedroom}).

A number of {\it house, floor} and {\it apartment} is realized by the predicate {\it  number}.

\subsubsection{Description of time}

A predicate of time takes the form:\\[10pt] 
{\it  time (cod\_tm, year, season, month, numb\_mon, day\_week, holyday, part\_day, hours, minutes)  }\\

\noindent where {\it  cod\_tm} is a code of time, {\it numb\_mon} is a date.

\subsubsection{Description of property}

A predicate of property is of the form:\\[10pt] 
{\it  property (cod\_prop, name,  scale,  state, cod\_obj)  }\\

\noindent where {\it  cod\_prop} is a code of property, {\it scale} is a name of scale, {\it state} is a state according to this scale, {\it cod\_obj} is a code of object to which this property belongs.

The estimation of properties is realized by means of scales\cite [sec.3-5]{Schank}. Scales can measure:
 
\begin{itemize}
\item  geometric parameters ({\it   length, volume, area});
\item  physical values ({\it   weight, temperature, charge, velocity, pressure});
\item  physical states ({\it hard, liquid, green, blue, cold, hot});
\item  psychological states ({\it   joy, distress, anger, fear});
\item  physiological characteristics ({\it roof pressure, pulse});
\item  economical activities ({\it income, expenses, profit});

\end{itemize}

\subsection{Typical predicates of sense relations}

Predicates of this type are divided into the next sorts: {\it  cause, relation, link}.

\subsubsection{Description of cause}

We consider causal dependence as physical or physiological regularity,  but no the result of simple induction \cite [part 6, ch.5]{RusselB}.
Besides, there is an inference as the kind of cause-and-effect relations. 

A predicate of cause takes the form:\\[10pt] 
{\it cause (cod\_cs, typ\_cs, typ\_sit, cod\_cause, cod\_res)  }\\

\noindent where {\it  cod\_cs}  is a code of cause, {\it  typ\_cs}  is a type of cause ({\it motive, objective causation, inference}),
{\it typ\_sit} is a type of situation ({\it event, action, thought, message}), {\it cod\_cause} is a code of cause, {\it cod\_res}  is a code of result.

\subsubsection{Description of sense relations}

The main role of predicate for sense relations is to map paradigmatic  relations \cite{Coboz}.
This predicate is of the form:\\[10pt] 
{\it relation (cod\_rel, typ\_rel, emot\_estim, first\_obj, sec\_obj, cod\_first, cod\_sec) }\\

\noindent where {\it  cod\_rel} is a code of relation, {\it  typ\_rel}  is a type of relation ({\it sexual, familiar, official, possession of something}), 
{\it emot\_estim} is an emotional estimation ({\it passion, love, animosity}), {\it first\_obj} denotes the first object of relation ({\it father, mother, brother, husband, chief}),
{\it sec\_obj} points to the second object of relation ({\it soon, daughter, sister, wife, subordinate}),
{\it cod\_first} is a code of the first object, {\it cod\_sec} is a code of the second object.

\subsubsection{Description of connections between assertions}

A corresponding predicate is of the form:\\[10pt] 
{\it link (cod\_link, base\_str, sub\_str, cod\_base, cod\_sub , conj, sem\_char) } \\

\noindent where {\it  cod\_link} is a code of connection, {\it base\_str} is a type of basic structure, {\it sub\_str} is a type of subordinate structure,
{\it cod\_base} is a code of basic structure,  {\it cod\_sub} is a code of subordinate structure, {\it conj} denotes a conjunction, {\it sem\_char} is a semantic characteristic of subordinate structure
({\it place, time, cause, purpose, condition, method}).

\subsection{Typical predicates of abstract objects}

Predicates of this type are divided into the next sorts: {\it  abstr, number}.

\subsubsection{Description of abstract objects of general kind}

A corresponding predicate takes the form:\\[10pt] 
{\it abstr (cod\_ab, concept, domain, ad\_prop, cod\_owner)  }\\

\noindent where {\it  cod\_ab} is a code of abstract object, {\it concept} denotes an appropriate concept, {\it domain} points to area of expertise ( {\it physics, mathematics, economics, sociology, control}), 
{\it ad\_prop} is additional information, {\it cod\_owner} is a code of owner (for {\it stock, deposits}) or author (for {\it articles, books, patents}).

\subsubsection{Description of numbers}

A predicate of number takes the form:\\[10pt] 
{\it  number (cod\_numb, descr\_word,  numb,  cod\_obj)}\\

\noindent where {\it  cod\_numb} is a code of number, {\it descr\_word} is a phrase describing this number, {\it numb} presents a decimal number, 
{\it cod\_obj} is a code of object connected with the given number.

\subsection{Typical predicates of actions and processes}

Predicates of this type are divided into the next sorts: {\it  action, process, thought, message}.

\subsubsection{Description of actions}

Actions of persons, organizations, machines, animals, natural phenomena are described using the predicate {\it  action}.
Such actions are usually localized  in a time and space. 
This predicate takes the form:\\[10pt] 
{\it  action (cod\_act, sem\_typ\_act, sort\_act, neg\_act, tense, char\_act, adverb, word, cod\_sub, cod\_obj, cod\_from\_obj, cod\_to\_obj, scale, res\_state, cod\_time, cod\_loc, cod\_way, cod\_purp, cod\_cause)}\\

\noindent where {\it  cod\_act} is a code of action, {\it sem\_typ\_act} is a semantic type of action (PROPEL, MOVE, GO, etc), {\it sort\_act} is a kind of action ({\it real, possible, necessary}), 
{\it neg\_act} denotes the negation of action, {\it tense} is a grammatical tense, {\it char\_act} is a character of action ({\it complete, incomplete}), {\it adverb} is an adverb which is used to estimate this action,
{\it word} is a verb describing this action, {\it cod\_sub} is a code of action subject, {\it cod\_obj} is a code of influence object, {\it cod\_from\_obj}  is a code of object from which the action is transferred, 
{\it cod\_to\_obj} is a code of object to which the action is directed, {\it scale} denotes a scale (see above),  {\it res\_state} is a state of subject or object (pointed by a value of scale) as a result of action, 
{\it cod\_time} is a code of action time, {\it cod\_loc} is a code of action place, {\it cod\_way} is a code of action way, {\it cod\_purp} is a code of action purpose, {\it cod\_cause} is a code of action cause.

\subsubsection{Description of processes}

Long procedures and other occupations of persons and  organizations are described by means of the predicate {\it  process} .
Such processes contain usually many actions. 
This predicate is of the form:\\[10pt] 
{\it  process (cod\_pr,  sort\_pr, typ\_pr, neg\_pr, tense, char\_pr, adverb, word, cod\_sub, cod\_obj,  cod\_start\_pr, cod\_end\_pr, cod\_start\_loc, cod\_end\_loc, cod\_way, cod\_purp, cod\_res)}\\

\noindent where {\it  cod\_pr} is a code of process, {\it sort\_pr} is a kind of process ({\it real, possible, necessary}), {\it typ\_pr} is a  type of process ({\it job, teaching, sport, hobby}),  
{\it neg\_pr} denotes the negation of process, {\it tense} is a grammatical tense, {\it char\_pr} is a character of process ({\it complete, incomplete}), {\it adverb} is an adverb which is used to estimate the process,
{\it word} is a verb describing this process, {\it cod\_sub} is a code of process subject, {\it cod\_obj} is a code of influence object,  {\it cod\_start\_pr} is a code of start time, {\it cod\_end\_pr} is a code of final time,
{\it cod\_start\_loc} is a code of start place, {\it cod\_end\_loc} is a code of final place,
{\it cod\_way} is a code of process way, {\it cod\_purp} is a code of process purpose, {\it cod\_res} is a code of process result.

\subsubsection{Description of thoughts}

Perceptions and thoughts of persons shaped by assertions are described using the predicate {\it  thought}.
This predicate takes the form:\\[10pt] 
{\it  thought (cod\_th,  sort\_th,  neg\_th, tense, char\_th, adverb, word, cod\_sub, cod\_obj,  cod\_time,  cod\_loc, cod\_purp)}\\

\noindent where {\it  cod\_th} is a code of thought, {\it sort\_th} is a kind of thought ({\it real, possible, necessary}),  
{\it neg\_th} denotes the negation of thought, {\it tense} is a grammatical tense, {\it char\_act} is a character of thought ({\it complete, incomplete}), {\it adverb} is an adverb which is used to estimate the thought,
{\it word} is a verb describing this thought, {\it cod\_sub} is a code of subject, {\it cod\_obj} are codes of thought sorts ({\it action, object, event, state}),
{\it cod\_time} is a code of time (when this thought arose), 
{\it cod\_loc} is a code of place (connected with this thought), {\it cod\_purp} is a code of solved problem.

\subsubsection{Description of message}

A  predicate of message is of the form:\\[10pt] 
{\it  message (cod\_ms,  sort\_ms,  neg\_ms, tense, char\_ms, adverb, word, cod\_sub, cod\_adr,  theme, cod\_time, cod\_loc, cod\_purp, cod\_way, cod\_cause)}\\

\noindent where {\it  cod\_ms} is a code of message, {\it sort\_ms} is a kind of message ({\it real, possible, necessary}),  
{\it neg\_ms} denotes the negation of message, {\it tense} is a grammatical tense, {\it char\_ms} is a character of message ({\it complete, incomplete}), {\it adverb} is an adverb which is used to estimate the message,
{\it word} is a verb describing this message, {\it cod\_sub} is a code of subject, {\it   cod\_adr} is a code of addressee,  {\it  theme} are codes of message sorts ({\it action, object, event, state }),
{\it cod\_time} is a code of message time, {\it cod\_loc} is a code of message place, {\it cod\_purp} is a code of message purpose,
{\it cod\_way} is a code of transfer way ({\it post, e-mail, verbally}), {\it cod\_cause} is a code of message cause.

\subsection{Typical predicates of states and events}

A change of {\it internal state} (for a physical object) is referred to as {\it an event}. 
States and events  of persons, organization, machines, things, animals, natural phenomena are described using the predicate {\it event}. 
This predicate takes the form:\\[10pt] 
{\it  event (cod\_evt, sort\_evt, neg\_evt, tense, char\_evt, adverb, word, cod\_sub, cod\_obj, scale, beg\_state, res\_state, cod\_time, cod\_loc, cod\_cause)}\\

\noindent where {\it  cod\_evt} is a code of event (state),  {\it sort\_evt} is a kind of event (state) ({\it real, possible, necessary}), 
{\it neg\_evt} denotes the negation of event (state), {\it tense} is a grammatical tense, {\it char\_evt} is a character of event (state) ({\it complete, incomplete}), {\it adverb} is an adverb which is used to estimate this event (state),
{\it word} is a verb describing this event (state), {\it cod\_sub} is a code of subject for this event (state), {\it cod\_obj} is a code of influence object, 
{\it scale} denotes a scale (see above),  {\it beg\_state} is a start state (pointed by a value of scale), {\it res\_state} is a final state, 
{\it cod\_time} is a code of time (for this event or state), {\it cod\_loc} is a code of  place (for this event or state), {\it cod\_cause} is a code of cause (for this event or state).
  
\section{A database}

A database contains a set of typical predicates describing facts (after the substitution of individual constants instead variables).
The database includes such predicates: {\it  person, organization, thing, machine, animal, abstr, nature, action, process, event, message, thought, property, number, place, time, cause, relation, link}. 
Besides, the database contains  dictionaries of paradigms, verbs and nouns\footnote{Described further project solutions are oriented to English. For other languages, these solutions  must be   modified.}.

\subsection{A dictionary of paradigms}
 
To represent a dictionary of paradigms, we use the predicate:\\[10pt]
{\it paradigm (cod\_par, osn\_form, paradigm, gram\_cat,  synt\_char)}\\

\noindent where   {\it  cod\_par} is a code of paradigm,  {\it  osn\_form} is a basic form of word,  {\it  gram\_cat} is a grammatical category,   {\it synt\_char} is a syntax characteristic.

A basic form is a basic word in dictionaries. It can be several descriptions with the same basic form to present all paradigms for all parts of speech.
A grammatical category points to belonging to a certain part of speech  ({\it  noun, verb, adverb}, etc). A syntax characteristic describes a syntax function of given word in a sentence.
For example, an adverb can be the next sorts: {\it  place, time, degree, manner}.

\subsection{A dictionary of nouns}

To represent a dictionary of nouns\footnote{Besides nouns, this dictionary contains as well adjectives.}, we use the predicate:\\[10pt]
{\it noun (cod\_int,  osn\_form, gram\_cat, maj\_class,  sem\_cod, verb, scale, state, combin )}\\

\noindent where {\it  cod\_int} is an unique code,  {\it  osn\_form} is a basic form of word, {\it gram\_cat} is a grammatical category (noun or adjective),  {\it maj\_class} is a generic class, {\it  sem\_cod} is a semantic code,  
{\it verb} is a corresponding verb for verbal noun, {\it scale} denotes a scale (see above),  {\it state} is a state pointed by a value of scale, {\it combin} describes the compatibility of nouns.

The variable {\it  sem\_cod} has the next values: {\it person}, {\it  prof} (profession), {\it  sibl} (sibling connection), {\it  org} (organization), {\it  anim}(animal), {\it plant, place, time, thing, weapon , mach} (machine), {\it  occup} (occupation),  {\it  event, state, prop} (property), 
{\it  body , natur} (natural phenomenon), {\it  mat} (matter), {\it  env} (environment), {\it psych} (psychic process or action), {\it  mes} (message), {\it  act} (physical action of person), {\it  abstr } (abstract concept), {\it  scale, cloth, food, quant } (attribute of quantity).

The description of  compatibility for nouns connected by means of  prepositions permits to point to admissible combinations of preposition and semantic  code (for example, for the word {\it book}: {\it on\_thing, on\_abstr, of\_person, of\_org}).

\subsection{A dictionary of verbs}

The predicate of this dictionary is of the form:\\[10pt]
{\it verb (cod\_verb,  inf, after\_verb, sem\_typ, scale,  beg\_state, end\_state, subj, contr\_at, contr\_from, contr\_to, contr\_with)}\\

\noindent where {\it  cod\_verb} is a code of verb, {\it inf} is a basic form (Infinitive), {\it  after\_verb} is an adverbial particle, {\it  sem\_typ} is a semantic type of verb (PROPEL, MOVE, GO, etc), 
{{\it beg\_state} is a start state (pointed by a value of scale) as a result of action, {\it end\_state} is a final state, {\it  subj} is semantic codes of subject (see the description of the predicate  {\it noun}),
{\it  contr\_at} is the description of control in the construction AT,  {\it  contr\_to} is the description of control in the construction TO, {\it  contr\_from} is the description of control in the construction FROM, 
{\it  contr\_with} is the description of control in the construction WITH. 

The description of control in the construction AT is used to point to semantic codes of  noun realizing a direct object (see the description of the predicate {\it noun}).
The control in the construction FROM characterizes  an indirect object from which an action is transferred. To do this, it is necessary to point  to admissible combinations of preposition and semantic  code (for example, for the verb {\it  get: from\_person, from\_org}).
The control in the construction TO presents  an indirect object to which an action is directed. To do this, it is necessary as well to point to admissible combinations of preposition and semantic  code (for example, for the verb {\it  put: on\_mach, on\_body, to\_thing}).
The control in the construction WITH describes possible tools and methods to realize the given action.

For example, for the verb  {\it  shoot} (in the meaning {\it kill}):

\begin{tabbing}
AAAAAAAA    \= AAAAAAAA    \kill
{\it  inf }        \> shoot         \\ 
{\it sem\_type}      \> PROPEL      \\
{\it scale}         \> HEALTH      \\ 
{\it beg\_state}    \> +50         \\
{\it end\_state}    \> -100         \\ 
{\it subj }  \> person      \\
{\it contr\_at }    \> person      \\
{\it contr\_with}     \> weapon      \\

\end{tabbing}

Depending on the sense, a verb can have several descriptions in the dictionary {\it verb}.

The dictionary maps the admissible compatibility of subject and objects with the character of action \footnote {The analogous approach  was proposed in \cite{Melchuk} using the concept of {\it semantic valence}.
The dependence of subject and objects on the type of verb is considered as well in works of Katz-Fodor \cite {Lyons}.}.
This permits to eliminate the semantic ambiguity for verbs(nouns) and to determine nonsensical sentences (see further).

\section{Semantic representations of sentences using typical predicates}

To represent every sentence, one must execute semantic descriptions:
 
\begin{itemize}
\item  groups of nouns;
\item  simple sentences;
\item  compound and complex sentences.
\end{itemize}  

The semantic representation  is based on syntax analysis of sentence. The description of grammar is executed by means of {\it  Backus-Naur form}\cite {Ostapov1} .

\subsection{Semantic representations of  noun groups }

Subjects, objects, adverbial modifiers are described  using {\it a group of noun}.
Consider {\it a simple group of noun} including a {\it basic noun} and {\it attributes}. At the beginning of group a preposition can be. 
If the basic noun describes a physical or abstract object, then this noun is transformed to the predicate of type: {\it  person, animal, organization, thing, machine, nature, abstr}. 
To determine the type of  predicate, the semantic code of this noun (pointed in {\it the dictionary of nouns}) is used. 
A verbal noun is considered in association with an action or event described in a given sentence(see further).

If the basic noun poses  a time or place, it is necessary to take account of a preposition. 
In this case,  the noun is transformed to the predicate of type: {\it time, place}.   

By means of article the individualizing character of basic noun is determined. If the attribute of basic noun is a property, then the predicate {\it  property} is formed for this attribute. 
If  the attribute is a participle, then the predicate of action or event is built. If nouns are used as attribute, then it can consider such group as an extended group of noun (see further).

When the group of noun is analysed, the  {\it  identification of objects}  is executed. The identification permits to establish the identity of objects. Algorithms of identification  are considered in  \cite {Ostapov2}.
If an identical object already exists for the given basic noun, then the new predicate is not formed, and only the available  predicate is modified.

To analyse {\it an extended group of noun} (connecting simple groups of nouns using prepositions), rules of compatibility for nouns are used (see {\it the dictionary of nouns}). 
If a noun denoting a property is connected with other noun using the preposition {\it of}, then the predicate {\it property} is formed for the first and pointed to this property in the second.
If the  preposition {\it of}  is used to present an owner of thing or machine, then this owner  is pointed  in the predicate {\it thing} or {\it machine}.
In other cases, rules of compatibility  are used. The connection between simple groups of nouns is realized with the predicate  {\it relation}.

\subsection{Semantic representations of simple sentences}

An analysed text is divided into fragments. Each fragment is characterized by unity of place and time. All actions or events of fragment are executed during a certain interval of time.
The place and time are described usually in the first sentences. According to the {\it principle of coherence} \cite[sec.9-2]{Lyons}, these place and time, as a rule, are not pointed  in following sentences.
Then to form the semantic representation for these sentences, the earlier--described  place and time  are used. 
Personal, demonstrative and possessive pronouns pose references to earlier--described actions, processes, events, states and objects.
Indefinite pronouns point to the existential quantifier ({\it  some, any}) and  the universal quantifier ({\it  all, every, each})  \cite [part II,ch.10]{RusselB}.

Consider the semantic representation of simple declarative sentence. After the semantics of a subject, objects, place, time, other adverbial modifiers is described, to represent the semantics of sentence,
it is necessary to form the semantic description of actions, processes, events, states. To do this, the analysis of predicate (in grammatical sense) is executed.
The following predicates are formed as a result of semantic analysis: {\it action} --- for physical actions, {\it message} --- for the transmission of information, 
{\it thought} --- for feelings and thoughts, {\it process} --- for long goal--seeking occupations, {\it event} --- for events and states. 
The choice of predicate is based on  the indication of semantic type in {\it the  dictionary of verbs}. 

Forming  of semantic predicates describing actions and events is reduced to the determination of factors which are typical for these actions and events: 
\begin{itemize}
\item subject of action (event);   
\item objects which take part in the transmission of information or action;
\item location and time of action (event);
\item purpose (result) and method (tool) of action (event).    
\end{itemize}  

Semantic roles of indirect objects and adverbial modifiers are determined using prepositions for noun groups.

To overcome the problem of ambiguous expressions, we select a semantic variant corresponding to one of  the descriptions of the given verb (in {\it the  dictionary of verbs}).
For example, consider the sentence  {\it  A man works with magnetic field}. As the verb {\it  work} is combined with the abstract object {\it field} using the preposition  {\it  with},
it can eliminate the meaning describing the place of action. 

Nonsensical  sentences are found as well using {\it the dictionary of verbs}. For instance, consider the sentence  {\it  A man plays physic}. 
The word  {\it physic} is inadmissible as the verb {\it play} is not combined with the object denoting  {\it  science}.

Paradigmatic relations between words of sentences ( see \cite{Coboz}) are described in a knowledge base and dictionaries of database.
For a concrete sentence, these relations are presented by means of  the predicate {\it relation}.
Syntagmatic relations are pointed using connections between actions (events) and subjects, objects, adverbial modifiers.
Such connections are formed by means of references in predicates {\it action, message, thought, process, event}(codes of  subject, objects, time, place, etc).
     
 \subsection{Semantic representations of  compound and complex  sentences}  
 
A complex declarative  sentence includes clauses, participle and infinitive constructions. For each clause and construction, the same algorithm is used as for a simple sentence.
The connection between the main sentence and subordinate constructions is realized with the predicate {\it link}.

A compound sentence consists of several simple  and complex sentences. The connection between independent sentences is described  with the predicate {\it link}.

\section{A knowledge base}

To form answers for logical questions, it is necessary to have the  dictionary containing explanations of words.
This dictionary is realized as the {\it knowledge base}\cite{Ostapov1,Ostapov3}.
The knowledge base consists of the predicates {\it  tperson, taction, torganization,...} which have precisely the same structure as  appropriate predicates of  database ({\it person, action, organization,...}).
Building the knowledge base is founded on  algorithms of syntax and semantic analysis described in \cite {Ostapov1}.

There is the set of words whose sense is implied in the system and realized by appropriate algorithms.
All  other words are explained by means of  knowledge base. These explanations pose practically  semantic(verbal) definitions of words.

The knowledge base includes the next descriptions: 

\begin{itemize}
\item general properties of physical and abstract objects;   
\item general properties of actions, processes, and events;
\item operations and scripts; 
\item schemas and plans.
  
\end{itemize}

  There is a special article (fragment) in the knowledge base for each description of concept. We limit ourselves to the description of general properties (for physical and abstract objects, actions, processes, events and operations).
The description of scripts, schemas, and plans is considered in \cite {Ostapov3}. Previously, we discuss the problem of primitives ---
the use of  minimum dictionary to explain words \cite[part II] {RusselB}\footnote{The problem of primitives is considered in \cite{Wierz}. 
Ch.K.Ogden had proposed the minimum dictionary of English including 850 words. If the control problem for complex social systems is considered  as the object domain,
then the correction of this dictionary is demanded.}.

The minimum dictionary contains such classes of words to present:

\begin{itemize}
\item types of objects ({\it person, organization, place, people, animal, plant,...});   
\item semantic role of words in sentence ({\it subject, object, way, tool, place, time, property, cause, purpose, ...}); 
\item basic actions, processes and events ({\it be, go, move,  message, transfer, create, attend, thought, consist, ...}); 
\item the direction of moving ({\it top, down, left, right});
\item scales ({\it color, length, weight,...});
\item abstract concepts ({\it part, whole, knowledge, decision, analysis,...});
\item physical concepts ({\it force, velocity,...});
\item mathematical concepts ({\it number, set, point, line, plane,...}).

\end{itemize}

It should be emphasized that primitives are words having monosemantic, intuitively obvious meaning.  
Such parts of speech as conjunction, preposition, pronoun, particle are included as well in the number of primitives.

To find primitives for a set of words S (describing an object domain), each definition will be presented as {\it Subject+Predicate}
where {\it Subject} is a determined word, {\it Predicate} contains a generic concept and difference.
If $A$ and $B$ are nonintersecting sets of words, then $A (B)$ means that  the set of words $A$  is determined by means of the set of words $B$.

\newtheorem{Def}{Definition}.

\begin{Def}

Let {\it S} is a set of words, $S_{0}, S_{1}, S_{2},...,S_{n}$ are subsets of $S$ so that 
$S_{i} \cap S_{j}=\oslash, i,j = 0,...,n, i\neq j,  S = S_{0} \cup S_{1} \cup S_{2} \cup...\cup S_{n}$.                                                      
Assume that $S_{1}(S_{0}), S_{2}(S_{0}\cup S_{1}),S_{3}(S_{0}\cup S_{1}\cup S_{2}),..., S_{n}(S_{0}\cup S_{1}\cup S_{2}...\cup S_{n-1}$).
Then $S_{0}$ is the set of primitives for $S$.

\end{Def}

It is easy to verify that if $S^{'} \supset S$, then for $S^{'}$ the set of primitives $S_{0}^{'}$  exists so that $S_{0} \subseteq S_{0}^{'}$.
Another way, each word in the object domain can be determined using primitives.

The general purpose of definitions is to explain the sense of words. This permits to solve logical questions, which demand the analysis of connections between concepts\cite {Ostapov3}.
It does not mean that we shall go the chain $S_{j1}, S_{j2}, S_{j3},...,S_{0}$ in the reverse order ($j1 >j2 >j3>...>0$ ) completely. For most calls to the knowledge base, it is sufficient to establish the conformity with available facts
\footnote{ For instance, it is necessary to check the availability of weapon for  a person. It is known that  this person has a pistol. The definition of pistol says that a pistol is a weapon which is kept in hand.
It is sufficient for the check.}.

\subsection{An article of noun}

An article of noun describes general properties of physical or abstract object.
To form the article of noun (for example, {\it doctor}), at first the phrase of this kind is  entered: 
{\it frame  is  a doctor}.   

Then functions of this noun are indicated:\\[10pt]
{\it doctor examines a person\\
doctor determines a disease\\
doctor prescribes a medicine}\\

The structure of object (for example, for the noun {\it car}) is formed by means of the phrase:\\[10pt]
{\it A car consists of chassis, engine,...}

\subsection{An article of verb}

An article of verb describes general properties of actions, processes, and events.
To form the article of verb (for example, {\it  to go on}), the phrase of type {\it frame is to go on} is entered.
To describe different values of verb,  the group of noun can be used in addition:\\[10pt]  
{\it frame  is to shoot from a gun\\     
frame  is to shoot a  person}\\    

Then semantic definitions and descriptions of concrete actions are entered.
 For example, after the phrase {\it frame is to learn}, there will be the descriptions:\\[10pt]  
{\it to do exercises\\
to answer a teacher\\
to visit a lesson in a class\\
to study a textbook}\\

After semantic definitions, motives and causes of action can be described.
For example, after the phrase  {\it frame is to shoot a  person}, there will be the expressions:\\[10pt]
{\it to kill the person by gun\\
to get money from the  person as a subject is criminal\\   
to pay off  the person as this person  outrages  a subject\\  
to annihilate the person as this person is the enemy of a subject}  

\subsection{An article of operation}

An operation is a sequence of actions to execute a certain purpose.
 The algorithm of  operation is based on principles of production systems \cite {Luger}.
 All conscious human activity includes the set of different operations  planed beforehand and modified in the course of realization.   

The description of operation can involve several alternatives.
Each alternative is entered with a separate article. 
For example, for the purpose {\it how to rob an organization} one can indicate the following description of the first alternative:\\[10pt]  
{\it frame is  how to rob an organization\\
alternative 1  ;  to go to the organization\\
alternative 1  ;  to come in\\
alternative 1  ;  to neutralize personal\\
alternative 1  ;  to open safes using tools\\
alternative 1  ;  to take moneys and things\\
alternative 1  ;  to come out}\\

The above sequence of actions  consists of stages. Each stage can contain an action, object of influence, motive and cause, condition and way (tool) of action.

\section{The comparison with other approaches to analysis of language}

Let us compare the proposed approach with other approaches to semantic analysis of language. We limit ourselves to the most advanced and widely known researches.

The most deep investigations of semantics (it is our opinion) are based on the representation of objects, events, processes and actions  in the view of {\it frames}
\cite{Minsk,Pospelov1,Pospelov2,Self}. Firstly, this approach is adequate ours in the sense that the representation of semantics (for objects, states, events, actions, and processes) is equivalent ours.
Secondly, for long-term saving of the great volume of information, it is necessary to use modern databases, i.e., we  arrive again at predicates.
Thirdly, the description of frame structures is a laborious process taking into account the great variety of objects, states, events, and actions.
The essential distinction of our approach implies that we use typical predicates and verbal definitions from encyclopedias.

The {\it conceptual approach} has gained wide acceptance \cite{Boris,Riesbeck,Schank,IPP,Birn}. The main imperfection of this approach is that objects, states, events, actions and processes have not adequate and detailed semantic description.
There is only the typification of action. This simplifies realization, but essentially constricts an area of solved tasks.  

The Russian school of semantics has gained essential successes\cite{Apr,Leont,Melchuk,Tusov}. The main feature of these works consists in the limitation of language aspects ---
paradigmatic and syntagmatic relations between words \cite[ch.1]{Coboz}. This does not permit to describe the functional purpose and structure of physical and abstract object, 
as well as  the functional role and complex character of action and processes. Of particular interest is the work \cite{Popov}. To describe facts and knowledge, the simple semantic net is applied --- each node is a word. 
This does not permit to discover in full measure the semantics of words and sentences. To present processes, in this work scripts were proposed.

\section{A formal logic system}

The computer system for natural language understanding can be considered as a formal logic system  $\mathcal{S}$ which is presented by the tuple:
\begin{equation}
    \mathcal{S} = < \mathfrak{A}, \mathfrak{B}, \Phi, \Sigma, \Omega >,                                                           
\end{equation}

\noindent where $\mathfrak{A}$ is an alphabet, $\mathfrak{B}$ is a set of atomic propositions,
$\Phi$ is a set of functions, $\Sigma$ are rules  of inference, $\Omega$ is a set of Horn's formulas (clauses).

The alphabet consists of letters, ciphers, special signs. Symbols are  formed from letters. Variables, individual constants, names of function and predicates are pointed using symbols.
A term is an individual constant, variable or expression $f(t_{1},t_{2},...,t_{n})$ where $t_{i}$ is a term, $f$ is a functional symbol.

{\it The atomic proposition} $P(t_{1},t_{2},...,t_{n})$ is formed from the predicate symbol $P$ and a list of terms. 
If terms do not contain variables, the atomic proposition is called {\it an atom}.

As well--formed formulas, only {\it Horn's clauses} are used\cite{Kow}:

\begin{equation}
  P_{j} \gets Q_{1}^{j},  Q_{2}^{j},..., Q_{n}^{j}, j=1,...,m,                                                           
\end{equation}

\noindent where $P_{j}$ and $Q_{i}^{j}$ are atomic propositions.

Atoms and Horn's clauses execute the function of axioms. Rules of inference use the {\it principle of resolution} and the {\it algorithm of unification} \cite{Kow}.
It can prove that every algorithm can be realized using this formal system.

The semantic interpretation of the formal system  $\mathcal{S}$ is the tuple:

\begin{equation}
    \mathfrak{I} = < \mathfrak{D}, I_{f},   I_{P}> ,                                                           
\end{equation}

\noindent where  $\mathfrak{D}$ is the {\it domain of interpretation} including ranges of variables for predicates and functions.
$\mathfrak{D}$  consists of sets of integer and real numbers, as well as sets of abstract words in the alphabet $\mathfrak{A}$. Each individual constant is connected with a certain value of variable.
$I_{f}$ is the map of sort:

\begin{equation}
    I_{f}(f) : M^{n} \to M
\end{equation}
 
\noindent for each function $f$ where $M$ is a set from   $\mathfrak{D}$.   $I_{P}$ is the map of sort:

\begin{equation}
    I_{P}(P) : M_{1}*M_{2}*...*M_{n} \to \lbrace true, false \rbrace
\end{equation}
 
\noindent for each predicate $P$ where $M_{i}$ is a range of  a variable with the index $i$ for this predicate. This range is included as well in  $\mathfrak{D}$.                                                    

Each concrete real object is characterized  by a compact connected domain of {\it physical space} at a given point of time and described by the above predicate $\psi$ (after the substitution of individual constants instead variables).
The {\it intension} of  $\psi$ is the aggregate of properties described by means of variables of this predicate, the {\it extension} of  $\psi$ is the set of real objects corresponding to this predicate \cite [ch.1]{Carnap}.
The above predicates $\varphi$ and $\gamma$ (after the substitution of individual constants instead variables) correspond to a set of compact connected domains in {\it a physical  space--time} (for real actions and events).
The predicates  $\psi$, $\varphi$ and $\gamma$  can use to describe possible objects, actions and events. For example, it is necessary for future time.

\begin{Def}

The aggregate  $\mathfrak{P} = \lbrace M_{ \psi}, M_{\varphi},  M_{\gamma}, M_{\delta},  M_{\mu}, M_{\rho}    \rbrace$ 
 is called  the semantic representation of a declarative sentence if  $M_{ \psi}$  is the set of predicates $\psi$ corresponding to physical objects from this sentence, $M_{\varphi}$ is the set of  predicates $\varphi$ presenting actions and processes of the given sentence,
$M_{\gamma}$ is the set of  predicates $\gamma$ describing events or states for objects of this sentence,   $M_{\mu}$ is the set of  predicates  $\mu$  corresponding to sense relations between objects from this sentence, 
$M_{\delta}$ is the set of predicates  $\delta$ corresponding to categories connected with  objects, actions and processes of the given sentence, $M_{\rho}$ is the set of predicates $\rho$ pointing to abstract objects from this sentence.
\end{Def}

For a concrete declarative sentence describing a real (i.e., checked) or possible (i.e., unchecked) fact, individual constants must be substituted instead variables in predicates $M_{ \psi}, M_{\varphi},  M_{\gamma}, M_{\delta},  M_{\mu}, M_{\rho}$. 
The {\it  extension} of $\mathfrak{P}$ is the value of truth for the given sentence \cite [ch.1]{Carnap}. 
If the sentence describes a real fact, then it is true. The {\it  intension} of  $\mathfrak{P}$ (in this case) is the real fact corresponding to a set of compact connected domains in {\it a physical space--time}.

Consider a general {\it synthetic sentence} presented by means of  $\mathfrak{P}$. If this sentence summarizes facts using the universal and existential quantifier, then the verification of a  such sentence can be a hard problem \cite [part 2, ch.10]{RusselB}. 
In this case, the {\it intension}  is the unification of facts.

The algorithm of forming  $\mathfrak{P}$  is realized in the system LEIBNIZ using the language PROLOG\cite{Ostapov1}. This language includes Horn's clauses and atomic propositions. 
An inference is executed with the help of resolution principle and unification algorithm. To check the accuracy of semantic representation, simple and logic questions are applied\cite{Ostapov2, Ostapov3}.

\section{Conclusion}

The proposed object--oriented approach poses the essential advance to solve the computer understanding problem. The representation of facts and knowledge using typical predicates permits greatly to simplify algorithms of processing and object descriptions.
Besides, the effectiveness of processing is increased. This is of great importance for the object domain having the large quantity of words. 
The main feature of our approach  (as compared with other investigations) is in the use of  encyclopedic knowledge. This  permits essentially  to extend the class of solved problems.

Using the proposed technology, new principles of control for complex social systems can be realized: both for enterprise  and  state structures.
For example, consider the problem of enterprise management.  At the moment  the technology ERP (Enterprise Resource Planning) is applied.
A new technology (in addition to ERP) is focused on  ensuring of profit increase and  normal cash flow\cite{Ostapov4,Ostapov5}.
This technology involves the next functions: forming  of intellectual interface on a natural language  to communicate with  a control system;
joint planning of production and sales to get the maximal profit; an adaptation of control system to internal and external events.

Other example concerns the domain of criminology.  Living computer systems save information about criminal offences by means of databases.
However, this information can not be used in full measure for logical processing as algorithms of semantic analysis are  not applied in such systems.
Thus, many actual problems are solved only by criminalists. Proposed algorithms permit to create  more intellectual  systems based on  methods of criminalistics.   
Such systems will  form answers for natural language questions about criminal offences using the purposeful selection from a database and logical processing of selected data.
If at first the great selection from the database is demanded, then it can gain serious results, which it can not get by other means.
Some algorithms of such task solution are considered in the paper\cite{Ostapov3}.

By this means, the presented technology discovers the perspective for the successful  solution of problems  in the social domains  with the help of intellectual computer systems.
We shall name this direction of investigations as {\it computer semantics}. Computer semantics is the part of cybernetics that studies questions connected with the application of  semantic representations (for expressions of natural language) to control problems.

\end{document}